\documentclass[runningheads]{llncs}
\usepackage{xcolor}
\usepackage[T1]{fontenc}
\usepackage{amsfonts,amsmath}
\usepackage{graphicx}
\usepackage{algorithm}
\usepackage{algorithmic}
\usepackage{tikz}
\usepackage{hyperref}
\usepackage{subcaption}
\usetikzlibrary{arrows.meta,positioning}

\begin{document}
\title{Dynamic Structural Causal Modeling for Sleep}
%
%
\author{Ranveer Singh\inst{1}\thanks{Equal Contribution} \and Saurabh Mathur\inst{2}$^{\star}$  \and Pranuthi Tenali\inst{1}  \and 
Arun Badi\inst{3} \and
Sriraam Natarajan\inst{1}}
\authorrunning{Singh et al.}
%
\institute{The University of Texas at Dallas, Richardson, USA \\
\email{\{ranveer.singh, pranuthi.tenali, sriraam.natarajan\}@utdallas.edu}\and
TU Darmstadt, Germany \\
\email{saurabh.mathur@tu-darmstadt.de} \and
ENT \& Sleep Medicine of Dallas, Dallas, USA\\
\email{arun.badi@gmail.com}}

\maketitle       
\begin{abstract}
The causal dynamics of sleep-disordered breathing are complex and vary across patient populations, hindering the development of targeted interventions. We learn dynamic causal graphs of sleep-disordered breathing from Home Sleep Apnea Test (HSAT) recordings, revealing systematic differences in causal structure across sex and age subcohorts. We do so using the PCMCI$+$ algorithm on windowed fractional variables derived from 105 HSAT recordings, exploiting domain knowledge via edge blacklisting and employing bootstrap aggregation to address small subcohort sizes. The learned graphs show that temporal self-dependencies and the apnea-desaturation relationship persist across all cohorts, while other relationships vary substantially. 

\end{abstract}

\section{Introduction}
Sleep-disordered breathing (SDB) is characterized by repeated disruptions to normal respiration during sleep, contributing to cardiovascular morbidity as well as metabolic dysfunction and daytime impairment~\cite{punjabi2009sleep}.
Moreover, sleep and SDB are dynamic; the respiratory and cardiovascular mechanisms underlying sleep vary throughout a single night and across nights, and the physiological mechanisms underlying SDB can differ among patients~\cite{eckert2013defining,eiseman2012impact,tschopp2021night}. These characteristics motivate the development of Dynamic Data-Driven Application Systems (DDDAS~\cite{darema2004dynamic}) for sleep 
that adapt according to the evolving physiological state of a patient during sleep.

However, such systems require a model of 
the causal mechanisms governing interactions between respiratory events in sleep.
Further, these mechanisms could differ across different subpopulations, potentially rendering a single population-level model inadequate~\cite{eckert2013defining}. Therefore, understanding the causal dynamics across different subpopulations is critical for developing reliable DDDAS systems for clinical decision support.

Learning such causal models for sleep is complicated by the high cost of polysomnography (PSG) tests, which are considered the gold standard. Home Sleep Apnea Tests (HSAT) offer a more accessible alternative, enabling patients to be tested at home at a lower cost and with less disruption. However, HSAT recordings have been used primarily for diagnostic scoring rather than for learning mechanistic models of sleep dynamics~\cite{cushman2026modified}.

To this end, we consider the task of identifying the differences in causal dynamics of sleep-disordered breathing across subcohorts stratified by age and sex. We do so by combining a small dataset of 105 HSAT recordings with domain knowledge in the form of potential causal relations and bootstrap aggregation to learn dynamic causal graphs. Specifically, we make the following key contributions: (1) we learn dynamic causal graphs of sleep-disordered breathing from HSAT recordings, showing that convenient at-home recordings can be used for causal modeling and serve as a component for a future DDDAS system, and (2) we perform a sex- and age-stratified dynamic causal analysis of SDB physiological signals, revealing meaningful structural differences across subpopulations, demonstrating the need to account for different subpopulations when developing causal models to serve as components for future DDDAS systems.

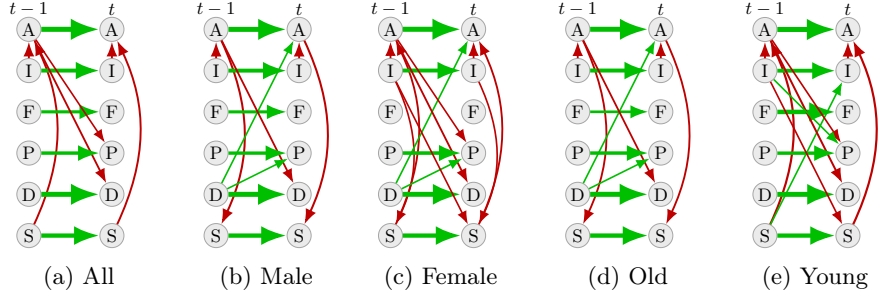
\begin{figure*}[!t]
    \centering
    \hfill
    \begin{subfigure}[t]{0.18\linewidth}
        \centering
        \resizebox{2cm}{!}{%
            \begin{tikzpicture}[
  >=Latex,
  varnode/.style={circle, draw=black!35, fill=black!8, minimum size=4.2mm, inner sep=0pt},
  positive/.style={green!75!black},
  negative/.style={red!75!black},
]
\node[font=\small] at (0.000,0.350) {$t-1$};
\node[font=\small] at (1.450,0.350) {$t$};
\node[varnode] (n_AHE_1) at (0.000,0.000) {A};
\node[varnode] (n_AHE_0) at (1.450,0.000) {A};
\node[varnode] (n_IFL_1) at (0.000,-0.720) {I};
\node[varnode] (n_IFL_0) at (1.450,-0.720) {I};
\node[varnode] (n_FLRE_1) at (0.000,-1.440) {F};
\node[varnode] (n_FLRE_0) at (1.450,-1.440) {F};
\node[varnode] (n_Pulse_1) at (0.000,-2.160) {P};
\node[varnode] (n_Pulse_0) at (1.450,-2.160) {P};
\node[varnode] (n_DeSat_1) at (0.000,-2.880) {D};
\node[varnode] (n_DeSat_0) at (1.450,-2.880) {D};
\node[varnode] (n_Snoring_1) at (0.000,-3.600) {S};
\node[varnode] (n_Snoring_0) at (1.450,-3.600) {S};
\draw[->,negative,line width=0.95pt] (n_IFL_1) to[bend left=0] (n_AHE_1);
\draw[->,negative,line width=0.95pt] (n_IFL_0) to[bend left=0] (n_AHE_0);
\draw[->,negative,line width=0.97pt] (n_Snoring_1) to[bend left=-25] (n_AHE_1);
\draw[->,negative,line width=0.97pt] (n_Snoring_0) to[bend left=-25] (n_AHE_0);
\draw[->,positive,line width=2.50pt] (n_AHE_1) -- (n_AHE_0);
\draw[->,positive,line width=1.54pt] (n_FLRE_1) -- (n_FLRE_0);
\draw[->,positive,line width=1.95pt] (n_IFL_1) -- (n_IFL_0);
\draw[->,negative,line width=0.93pt] (n_AHE_1) -- (n_DeSat_0);
\draw[->,positive,line width=2.48pt] (n_DeSat_1) -- (n_DeSat_0);
\draw[->,negative,line width=0.81pt] (n_AHE_1) -- (n_Pulse_0);
\draw[->,positive,line width=1.70pt] (n_Pulse_1) -- (n_Pulse_0);
\draw[->,positive,line width=2.14pt] (n_Snoring_1) -- (n_Snoring_0);
\end{tikzpicture}
        }
        \caption{All}
    \end{subfigure}%
    \hfill
    \begin{subfigure}[t]{0.18\linewidth}
        \centering
        \resizebox{2cm}{!}{%
            \begin{tikzpicture}[
  >=Latex,
  varnode/.style={circle, draw=black!35, fill=black!8, minimum size=4.2mm, inner sep=0pt},
  positive/.style={green!75!black},
  negative/.style={red!75!black},
]
\node[font=\small] at (0.000,0.350) {$t-1$};
\node[font=\small] at (1.450,0.350) {$t$};
\node[varnode] (n_AHE_1) at (0.000,0.000) {A};
\node[varnode] (n_AHE_0) at (1.450,0.000) {A};
\node[varnode] (n_IFL_1) at (0.000,-0.720) {I};
\node[varnode] (n_IFL_0) at (1.450,-0.720) {I};
\node[varnode] (n_FLRE_1) at (0.000,-1.440) {F};
\node[varnode] (n_FLRE_0) at (1.450,-1.440) {F};
\node[varnode] (n_Pulse_1) at (0.000,-2.160) {P};
\node[varnode] (n_Pulse_0) at (1.450,-2.160) {P};
\node[varnode] (n_DeSat_1) at (0.000,-2.880) {D};
\node[varnode] (n_DeSat_0) at (1.450,-2.880) {D};
\node[varnode] (n_Snoring_1) at (0.000,-3.600) {S};
\node[varnode] (n_Snoring_0) at (1.450,-3.600) {S};
\draw[->,negative,line width=0.95pt] (n_IFL_1) to[bend left=0] (n_AHE_1);
\draw[->,negative,line width=0.95pt] (n_IFL_0) to[bend left=0] (n_AHE_0);
\draw[->,negative,line width=1.01pt] (n_AHE_1) to[bend left=25] (n_Snoring_1);
\draw[->,negative,line width=1.01pt] (n_AHE_0) to[bend left=25] (n_Snoring_0);
\draw[->,positive,line width=2.48pt] (n_AHE_1) -- (n_AHE_0);
\draw[->,positive,line width=0.81pt] (n_DeSat_1) -- (n_AHE_0);
\draw[->,positive,line width=1.52pt] (n_FLRE_1) -- (n_FLRE_0);
\draw[->,positive,line width=1.95pt] (n_IFL_1) -- (n_IFL_0);
\draw[->,negative,line width=0.91pt] (n_AHE_1) -- (n_DeSat_0);
\draw[->,positive,line width=2.50pt] (n_DeSat_1) -- (n_DeSat_0);
\draw[->,positive,line width=0.80pt] (n_DeSat_1) -- (n_Pulse_0);
\draw[->,positive,line width=1.66pt] (n_Pulse_1) -- (n_Pulse_0);
\draw[->,positive,line width=2.15pt] (n_Snoring_1) -- (n_Snoring_0);
\end{tikzpicture}%
        }
        \caption{Male}
    \end{subfigure}
    \begin{subfigure}[t]{0.18\linewidth}
        \centering
        \resizebox{2cm}{!}{%
            \begin{tikzpicture}[
  >=Latex,
  varnode/.style={circle, draw=black!35, fill=black!8, minimum size=4.2mm, inner sep=0pt},
  positive/.style={green!75!black},
  negative/.style={red!75!black},
]
\node[font=\small] at (0.000,0.350) {$t-1$};
\node[font=\small] at (1.450,0.350) {$t$};
\node[varnode] (n_AHE_1) at (0.000,0.000) {A};
\node[varnode] (n_AHE_0) at (1.450,0.000) {A};
\node[varnode] (n_IFL_1) at (0.000,-0.720) {I};
\node[varnode] (n_IFL_0) at (1.450,-0.720) {I};
\node[varnode] (n_FLRE_1) at (0.000,-1.440) {F};
\node[varnode] (n_FLRE_0) at (1.450,-1.440) {F};
\node[varnode] (n_Pulse_1) at (0.000,-2.160) {P};
\node[varnode] (n_Pulse_0) at (1.450,-2.160) {P};
\node[varnode] (n_DeSat_1) at (0.000,-2.880) {D};
\node[varnode] (n_DeSat_0) at (1.450,-2.880) {D};
\node[varnode] (n_Snoring_1) at (0.000,-3.600) {S};
\node[varnode] (n_Snoring_0) at (1.450,-3.600) {S};
\draw[->,negative,line width=0.90pt] (n_IFL_1) to[bend left=0] (n_AHE_1);
\draw[->,negative,line width=0.90pt] (n_IFL_0) to[bend left=0] (n_AHE_0);
\draw[->,negative,line width=0.93pt] (n_Snoring_1) to[bend left=-25] (n_AHE_1);
\draw[->,negative,line width=0.93pt] (n_Snoring_0) to[bend left=-25] (n_AHE_0);
\draw[->,negative,line width=0.83pt] (n_IFL_1) to[bend left=25] (n_Snoring_1);
\draw[->,negative,line width=0.83pt] (n_IFL_0) to[bend left=25] (n_Snoring_0);
\draw[->,positive,line width=2.50pt] (n_AHE_1) -- (n_AHE_0);
\draw[->,positive,line width=0.82pt] (n_DeSat_1) -- (n_AHE_0);
\draw[->,positive,line width=1.94pt] (n_IFL_1) -- (n_IFL_0);
\draw[->,negative,line width=0.93pt] (n_AHE_1) -- (n_DeSat_0);
\draw[->,positive,line width=2.45pt] (n_DeSat_1) -- (n_DeSat_0);
\draw[->,negative,line width=0.81pt] (n_AHE_1) -- (n_Pulse_0);
\draw[->,positive,line width=0.82pt] (n_DeSat_1) -- (n_Pulse_0);
\draw[->,positive,line width=1.78pt] (n_Pulse_1) -- (n_Pulse_0);
\draw[->,negative,line width=0.81pt] (n_IFL_1) -- (n_Snoring_0);
\draw[->,positive,line width=2.13pt] (n_Snoring_1) -- (n_Snoring_0);
\end{tikzpicture}%
        }
        \caption{Female}
    \end{subfigure}%
    \hfill
    \begin{subfigure}[t]{0.18\linewidth}
        \centering
        \resizebox{2cm}{!}{%
            \begin{tikzpicture}[
  >=Latex,
  varnode/.style={circle, draw=black!35, fill=black!8, minimum size=4.2mm, inner sep=0pt},
  positive/.style={green!75!black},
  negative/.style={red!75!black},
]
\node[font=\small] at (0.000,0.350) {$t-1$};
\node[font=\small] at (1.450,0.350) {$t$};
\node[varnode] (n_AHE_1) at (0.000,0.000) {A};
\node[varnode] (n_AHE_0) at (1.450,0.000) {A};
\node[varnode] (n_IFL_1) at (0.000,-0.720) {I};
\node[varnode] (n_IFL_0) at (1.450,-0.720) {I};
\node[varnode] (n_FLRE_1) at (0.000,-1.440) {F};
\node[varnode] (n_FLRE_0) at (1.450,-1.440) {F};
\node[varnode] (n_Pulse_1) at (0.000,-2.160) {P};
\node[varnode] (n_Pulse_0) at (1.450,-2.160) {P};
\node[varnode] (n_DeSat_1) at (0.000,-2.880) {D};
\node[varnode] (n_DeSat_0) at (1.450,-2.880) {D};
\node[varnode] (n_Snoring_1) at (0.000,-3.600) {S};
\node[varnode] (n_Snoring_0) at (1.450,-3.600) {S};
\draw[->,negative,line width=0.91pt] (n_IFL_1)-- (n_AHE_1);
\draw[->,negative,line width=0.91pt] (n_IFL_0) -- (n_AHE_0);
\draw[->,negative,line width=0.94pt] (n_AHE_1) to[bend left=25] (n_Snoring_1);
\draw[->,negative,line width=0.94pt] (n_AHE_0) to[bend left=25] (n_Snoring_0);
\draw[->,positive,line width=2.47pt] (n_AHE_1) -- (n_AHE_0);
\draw[->,positive,line width=0.82pt] (n_DeSat_1) -- (n_AHE_0);
\draw[->,positive,line width=1.31pt] (n_FLRE_1) -- (n_FLRE_0);
\draw[->,positive,line width=1.97pt] (n_IFL_1) -- (n_IFL_0);
\draw[->,negative,line width=0.90pt] (n_AHE_1) -- (n_DeSat_0);
\draw[->,positive,line width=2.50pt] (n_DeSat_1) -- (n_DeSat_0);
\draw[->,positive,line width=0.81pt] (n_DeSat_1) -- (n_Pulse_0);
\draw[->,positive,line width=1.71pt] (n_Pulse_1) -- (n_Pulse_0);
\draw[->,positive,line width=2.16pt] (n_Snoring_1) -- (n_Snoring_0);
\end{tikzpicture}%
        }
        \caption{Old}
    \end{subfigure}%
    \hfill
    \begin{subfigure}[t]{0.18\linewidth}
        \centering
        \resizebox{2cm}{!}{%
            \begin{tikzpicture}[
  >=Latex,
  varnode/.style={circle, draw=black!35, fill=black!8, minimum size=4.2mm, inner sep=0pt},
  positive/.style={green!75!black},
  negative/.style={red!75!black},
]
\node[font=\small] at (0.000,0.350) {$t-1$};
\node[font=\small] at (1.450,0.350) {$t$};
\node[varnode] (n_AHE_1) at (0.000,0.000) {A};
\node[varnode] (n_AHE_0) at (1.450,0.000) {A};
\node[varnode] (n_IFL_1) at (0.000,-0.720) {I};
\node[varnode] (n_IFL_0) at (1.450,-0.720) {I};
\node[varnode] (n_FLRE_1) at (0.000,-1.440) {F};
\node[varnode] (n_FLRE_0) at (1.450,-1.440) {F};
\node[varnode] (n_Pulse_1) at (0.000,-2.160) {P};
\node[varnode] (n_Pulse_0) at (1.450,-2.160) {P};
\node[varnode] (n_DeSat_1) at (0.000,-2.880) {D};
\node[varnode] (n_DeSat_0) at (1.450,-2.880) {D};
\node[varnode] (n_Snoring_1) at (0.000,-3.600) {S};
\node[varnode] (n_Snoring_0) at (1.450,-3.600) {S};
\draw[->,negative,line width=0.95pt] (n_IFL_1) to[bend left=0] (n_AHE_1);
\draw[->,negative,line width=0.95pt] (n_IFL_0) to[bend left=0] (n_AHE_0);
\draw[->,negative,line width=1.08pt] (n_Snoring_1) to[bend left=-25] (n_AHE_1);
\draw[->,negative,line width=1.08pt] (n_Snoring_0) to[bend left=-25] (n_AHE_0);
\draw[->,positive,line width=2.50pt] (n_AHE_1) -- (n_AHE_0);
\draw[->,positive,line width=2.24pt] (n_FLRE_1) -- (n_FLRE_0);
\draw[->,positive,line width=1.91pt] (n_IFL_1) -- (n_IFL_0);
\draw[->,positive,line width=0.81pt] (n_Snoring_1) -- (n_IFL_0);
\draw[->,negative,line width=0.99pt] (n_AHE_1) -- (n_DeSat_0);
\draw[->,positive,line width=2.46pt] (n_DeSat_1) -- (n_DeSat_0);
\draw[->,negative,line width=0.83pt] (n_AHE_1) -- (n_Pulse_0);
\draw[->,positive,line width=0.82pt] (n_IFL_1) -- (n_Pulse_0);
\draw[->,positive,line width=1.69pt] (n_Pulse_1) -- (n_Pulse_0);
\draw[->,negative,line width=0.82pt] (n_IFL_1) -- (n_Snoring_0);
\draw[->,positive,line width=2.12pt] (n_Snoring_1) -- (n_Snoring_0);
\end{tikzpicture}%
        }
        \caption{Young}
    \end{subfigure}
    \caption{The Average Time Series Causal Graph across the different subcohorts and the entire cohort learnt using PCMCI$+$. The color of the edges indicates the sign of the Momentary Conditional Independence (MCI), with green being positive and red being negative. The thickness of the edge indicates the magnitude of MCI, with greater thickness corresponding to a larger magnitude.}
    \label{fig:graphs_learnt}
\end{figure*}
\section{Background}
The problem of learning dynamic causal models for reliable Dynamic Data-Driven Application Systems is related to the following areas.

\subsection{Dynamic Structural Causal Models}

A Dynamic Structural Causal Model (DSCM~\cite{boeken2024dynamicstructuralcausalmodels}) is a model for learning temporal causal dynamics. It is an extension of a structural causal model or an SCM, which is a mathematical framework for modeling the underlying data-generating process by capturing the causal relationships among the variables in a system~\cite{pearl2009causality}. An SCM, $\mathcal{M}$, is descibed as a tuple $\langle\mathbf{U}, \mathbf{V},  \mathcal{F}, P(\mathbf{U})\rangle$ where $\mathbf{U}$ is a set of exogenous variables, $\mathbf{V}$ is the set of endogenous variables determined by other variables in the model $\mathbf{U} \cup \mathbf{V}$, $\mathcal{F}$ is a a set of structural equations $f_i: U_i, V_{Pa_i} \rightarrow V_i$ where $Pa_i \subseteq \mathbf{V}\backslash V_i$, and $P(\mathbf{U})$ is a probability distribution~\cite{bareinboim2025causal}.
The structural equations of an SCM induce a graph $G = (\mathbf{V}, \mathbf{E})$ where an edge $X \rightarrow Y \in \mathbf{E}$ iff $X \in Pa_Y$.

A DSCM~\cite{boeken2024dynamicstructuralcausalmodels} extends SCMs to temporally evolving processes. Formally, a DSCM is defined $\mathcal{M}_d = \langle \mathbf{U}, \mathbf{V}, \mathcal{T}, \mathcal{F},  \mathbf{P(U)}\rangle$ where $\mathbf{U}$ and $\mathbf{V}$ are now sets of time indexed exogeneous and endogeneous processes respectively, $\mathcal{T} \subseteq \mathbb{R}$ is the time domain of the processes, $\mathcal{F}$ is a set of structural equations $f_i: V_{Pa_i}, U_i \rightarrow V_i$ mapping to endogeneous processes in $\mathbf{V}$  with temporal causality enforcing that $V_i^t$ depends only on values at time $t' \leq t $, and $P(\mathbf{U})$ is a probability measure over the trajectory of exogenous processes.

The structural equations $\mathcal{F}$ and probability distribution P$(\mathbf{U})$ of DSCMs generally cannot be identified from observational data alone, which is often the only type of data available. Therefore, we may need to make assumptions such as the causal Markov condition (each variable is independent of its non-descendants given its parents), faithfulness (the distribution does not entail additional independencies beyond the graph), causal sufficiency (there are no hidden confounders), and stationarity (the distribution of the process doesn't change over time), to learn a causal graph induced by a DSCM. When time is discretized and sampled, a time-series graph over lagged and contemporaneous dependencies is induced. Formally, we define such causal graphs as  $G = (\mathbf{V}, \mathbf{E}, \mathcal{T})$ where $\mathcal{T}$ is the discrete time domain, $\mathbf{V}$ is a set of time indexed observed variables with $V_i^t$ indicating value of variable $V_i$ at time $t \in \mathcal{T}$, and $\mathbf{E}$ is a set of edges which are either contemporous $V_i^t \rightarrow V_j^t$ where $j \neq i$, or lagged $V_i^{t-\tau} \rightarrow V_k^{t}$ where $\tau> 0$. These graphs can be learned using algorithms such as PCMCI$+$~\cite{pcmci_plus}

\begin{figure}[!t]
    \centering
    \includegraphics[width=0.4\linewidth]{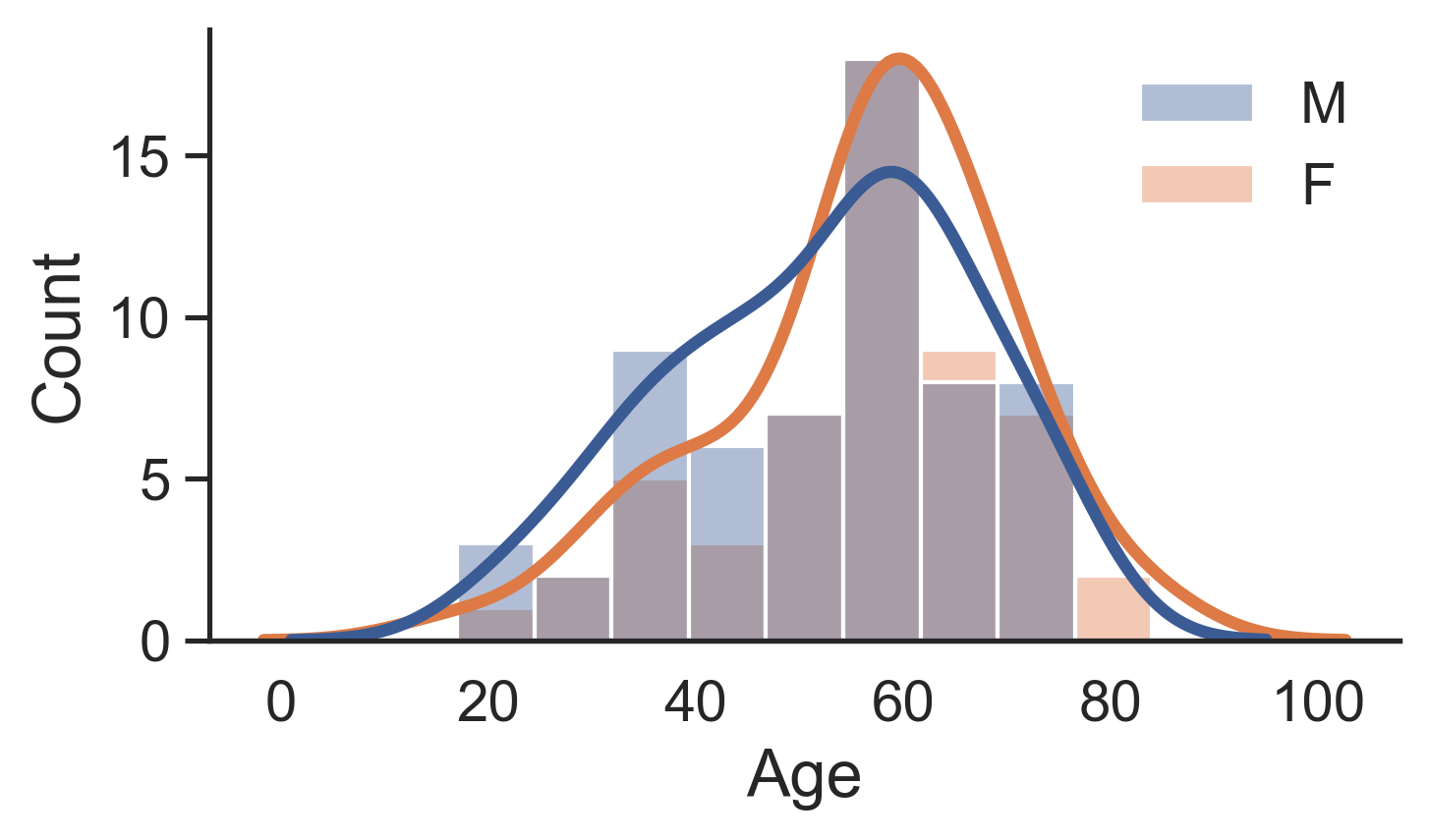}
    \caption{Age distribution for the patients based on sex}
    \label{fig:age_distribution}
\end{figure}
\subsection{Domain Knowledge as Constraints}
Even though it is possible to learn causal graphs from observational data, these models will only be learned up to a Markov Equivalence class~\cite{pearl2009causality}, i.e, a set of graphs learned from the observational data with many edge directions (direction of causality) unresolved. Therefore, background knowledge such as forbidden and required edges~\cite{meek1995causal} provided either by a domain expert or a Large Language Model (LLM)~\cite{2025mathurAIME} is essential to orient such edges and learn an unambiguous causal graph. Additionally, in a temporal setting, domain knowledge about timescales, known causal mechanisms, and contemporaneous relationships among features can be used to learn a causal model~\cite{runge2019detecting}.
\subsection{Discovering Higher-Order Interaction}
Causal discovery is difficult in domains such as medicine, where acquiring interventional data is difficult or unethical, and observational data is scarce due to the high cost of acquisition~\cite{liu2026causal}. In such a low-data setting, the causal graphs learned are highly unstable and susceptible to noise. One way to mitigate this issue is through data bootstrapping, where data is sampled with replacement, and the resulting graphs are aggregated using techniques such as total weight averaging (TWA) to generate a final causal graph with confidence estimates for the causal edges~\cite{debeire2024bootstrap,friedman1999,SCUTARI2013207}.  However, in a complex domain such as medicine, where multiple variables act together, aggregation that only considers the edges is insufficient. Local structures such as colliders, forks, and chains are more stable and meaningful, and aggregating across such structures can allow us to learn better and more stable causal graphs for such domains~\cite{zanga2025causal}.

\section{Learning Causal Dynamics for Sleep}
We consider the task of learning causal dynamics for sleep using the dataset of HSAT recordings. We formalize the task as follows
\begin{center}
\fbox{
\parbox{0.95\linewidth}{
\textbf{Given:} A temporal dataset $\mathcal{D}$ over $\{X_1,X_2,\dots,X_n\}$, where $X_i^j(t)$ denotes the value of feature $j$ for patient $i$ at time $t,$ and domain knowledge $\mathcal{K}$ in the form of the set of forbidden causal edges.
\\
\textbf{To Do:} Learn a time-series causal graph $\mathcal{G}$ modelling the contemporaneous and lagged dependencies between variables derived from the data.
}
}
\end{center}
In the subsequent subsections, we present the construction of the windowed dataset over which the model will be learned, followed by model construction and comparison of the models across different subcohorts\footnote[1]{The experimental setup, feature definition and construction, and code are provided in the supplementary at \url{https://github.com/s-ranveer/causal_dynamics_sleep}}.
\subsection{Dataset}
\label{dataset}
We consider the HSAT dataset provided by our clinical expert.
The dataset provided consisted of 105 recordings, each at least 2 hours long, with the following features: Snoring, Pulse, Oxygen Saturation, Effort, and Flow. The patients are split 58 males to 47 females based on sex, with the overall age distribution shown in Figure~\ref{fig:age_distribution}. Furthermore, males aged under 40 and females aged under 50 are considered young; otherwise, they are considered old. To learn causal models, we constructed a windowed dataset with fractional variables computed over 10-second windows~\cite{berry2012rules}. The variables are presented
in Table~\ref{tab:features} with the average values\footnote[2]{The zero value for windowed feature \textbf{F} might be specific to the current cohort and approximations used.} across the different subcohorts shown in Table~\ref{tab:fractions}.
\begin{table}[!t]
    \centering
    \caption{Constructed Windowed Features as a fraction of the analysis window}
    \begin{tabular}{|l|l|}\hline
         \textbf{Feature} & \textbf{\thinspace\quad\quad\quad\quad \quad Description} \\
         & \textbf{\quad\thinspace (Fraction of the analysis window)} \\\hline
         Apnea Hypopnea Events (A)& Occupied by Apnea and Hypopnea events~\cite{malhotra2024aasm}\\
         Desaturation (D) & Spent in Oxygen Desaturation episodes~\cite{temirbekov2018ignored}\\
         Flow Limitation (I) &  Exhibiting inspiratory flow-limited breathing~\cite{guevarra2022immediate}\\
         Flow Limited Respiratory Effort (F)  & Characterized by Flow-limited breathing\\
          & with elevated respiratory effort\\
         Pulse Activation (P) & Transient pulse-rate elevations \\
         Snoring Burden (S) & Sustained snoring activity~\cite{maimon2010does} \\         
    \hline
    \end{tabular}

    \label{tab:features}
\end{table}
\begin{table}[!t]
    \centering
    \caption{Average value of the fractional features across the different cohorts}
    \begin{tabular}{|l|c|c|c|c|c|}\hline
         \textbf{Feature} & \textbf{All} & \textbf{Male} & \textbf{Female} & \textbf{Old} & \textbf{Young}  \\\hline
         A & 0.02$\scriptstyle\pm 0.06$ & $0.02 \scriptstyle\pm 0.06$ & $0.02 \scriptstyle\pm 0.06$ & $0.02 \scriptstyle\pm 0.05$ & $0.03 \scriptstyle\pm 0.07$ \\
         D & $0.02 \scriptstyle\pm 0.02$ & $0.02 \scriptstyle\pm 0.03$ & $0.01 \scriptstyle\pm 0.02$ & $0.01 \scriptstyle\pm 0.02$ & $0.02 \scriptstyle\pm 0.03$ \\
         F ($\times 10^{-5}$) & $1.36 \scriptstyle\pm 7.47$ & $2.43 \scriptstyle\pm 9.89$ & $0.00 \scriptstyle\pm 0.00$ & $1.30 \scriptstyle\pm 7.61$ & $1.61 \scriptstyle\pm 7.36$ \\
         I & $0.29 \scriptstyle\pm 0.12$ & $0.28 \scriptstyle\pm 0.13$ & $0.30 \scriptstyle\pm 0.10$ & $0.30 \scriptstyle\pm 0.11$ & $0.24 \scriptstyle\pm 0.12$ \\
         P & $0.03 \scriptstyle\pm 0.03$ & $0.04 \scriptstyle\pm 0.03$ & $0.03 \scriptstyle\pm 0.02$ & $0.03 \scriptstyle\pm 0.03$ & $0.03 \scriptstyle\pm 0.03$ \\
         S & $0.25 \scriptstyle\pm 0.14$ & $0.25 \scriptstyle\pm 0.12$ & $0.26 \scriptstyle\pm 0.16$ & $0.25 \scriptstyle\pm 0.14$ & $0.25 \scriptstyle\pm 0.15$ \\\hline
    \end{tabular}
    \label{tab:fractions}
\end{table}
\begin{algorithm}[!t]
\caption{Learning Causal Dynamics with Aggregation over Higher-Order Structures}
\label{alg:pseudocode}
\begin{algorithmic}
\REQUIRE Dataset $\mathcal{D}$, prior knowledge $\mathcal{K}$, number of bootstraps $b$, threshold $\alpha$, MCI threshold $m$, maximum temporal lag $\tau_\text{max}$
\ENSURE The final causal graph $\mathcal{G}^*$

\textbf{// Step 1: Bootstrapped Time Series Causal Discovery}
\STATE Initialize the set of graphs $\mathbf{G} = \{\}$
\FORALL{$i \in 1$ to $b$}
    \STATE $\mathcal{D}_i \gets$ Sample $|\mathcal{D}|$ patient trajectories with replacement from $\mathcal{D}$
    \STATE $\mathcal{G}_i \gets$ \textsc{LearnTimeSeriesGraph}$(\mathcal{D}_i, \mathcal{K}, \tau_\text{max}, m)$
    \STATE $\mathbf{G} \gets \mathbf{G} \cup \{\mathcal{G}_i\}$
\ENDFOR

\medskip
\textbf{// Step 2: Extract Higher-Order Structures}
\STATE Initialize structure count map $\Phi = \{\}$
\FORALL{$\mathcal{G}_i \in \mathbf{G}$}
    \FORALL{$V_j \in \mathbf{V}$}
        \STATE $E_{\phi_{V_j}} \gets \{e_{V_k V_j} \mid V_k \in \Pi_{V_j}^i\}$ \COMMENT{//Edge set incident on $V_j$ from its parents in $\mathcal{G}_i$}
        \STATE $\phi_{V_j} \gets (V_j,\ E_{\phi_{V_j}})$
        \STATE $\Phi[\phi_{V_j}] \gets \Phi[\phi_{V_j}] + 1$
    \ENDFOR
\ENDFOR

\medskip
\textbf{// Step 3: Generalized Model Averaging (GMA)}
\STATE Normalize $\Phi$ to obtain local posterior estimates:
$P(\phi_{V_i} \mid \mathcal{D}) = \frac{\Phi[\phi_{V_i}]}{\sum_{\phi} \Phi[\phi]}$
\STATE $\tilde{\Phi} \gets \{\phi_{V_i} \mid P(\phi_{V_i} \mid \mathcal{D}) > \alpha\}$
\COMMENT{// Apply threshold}
\STATE Sort $\tilde{\Phi}$ by decreasing $P(\phi_{V_i} \mid \mathcal{D})$

\medskip
\textbf{// Step 4: Greedy Acyclic Graph Construction}
\STATE $\mathcal{G}^* \gets (\mathbf{V}, \emptyset)$
\FORALL{$\phi_{V_i} \in \tilde{\Phi}$}
    \STATE $\mathcal{G}' \gets (\mathbf{V},\ E_{\mathcal{G}^*} \cup E_{\phi_{V_i}})$
    \IF{\textsc{IsAcyclic}$(\mathcal{G}')$}
        \STATE $\mathcal{G}^* \gets \mathcal{G}'$
    \ENDIF
\ENDFOR
\RETURN $\mathcal{G}^*$

\end{algorithmic}
\end{algorithm}
\subsection{Model Construction}
\label{algo}
To learn causal dynamics for sleep, we consider a framework that combines time-series causal discovery with aggregation over higher-order local causal structures to handle the noise inherent in small data settings and to learn stable, more meaningful causal graphs.

Algorithm~\ref{alg:pseudocode} describes the overall pipeline for the task. First, we sample patient trajectories from the dataset with replacement $b$ times. We then use the PCMCI$+$ algorithm with a partial correlation test to learn initial time-series causal graphs for each bootstrap, considering only edges consistent with the background knowledge $\mathcal{K}$, removing any edges with a Momentary Conditional Independence (MCI~\cite{runge2019detecting}) below the threshold $m$, to obtain the graphs $\mathbf{G}$. Next, we construct the final graph using General Model Averaging (GMA~\cite{zanga2025causal}). For each graph, we obtain the higher-order structure by selecting all subgraphs induced by each vertex and its lagged and contemporaneous edges with its parents $(V_j, E_{\phi_{V_j}})$. We compute the posterior over these structures as their normalized counts over the different bootstraps. We remove any structures with a posterior value below the threshold $\alpha$, and get a sorted list of the structures in descending order of their posterior value. We construct the final dynamic causal graph by adding edges for each structure in the sorted list in order, skipping any structure if it violates the acyclicity constraint. 
\subsection{Results}

We consider the different subcohorts from the dataset defined in \ref{dataset}, and the algorithm defined in \ref{algo}, using PCMCI$+$~\cite{pcmci_plus} as the underlying time-series causal learning algorithm. We elicited background knowledge in the form of forbidden and allowed edges from a domain expert and incorporated it as constraints into the causal discovery process. The parameters $\tau_\text{max} = 1, \alpha=\frac{1}{|V| - 1}  = 0.09, b = 100$, and $m = 0.03$ were either provided by the domain expert or adopted from the original works~\cite{Liao2022On,zanga2025causal}. The different time-series causal graphs learnt across the different subcohorts are shown in Figure~\ref{fig:graphs_learnt}.

When considering the subcohorts based on sex, one difference between the male and female subcohorts is the relationship between Snoring (S) and Inspiratory Flow Limit (I) in the same window, with such a relationship present in females but not in males. Specifically, the structure $\text{S}_{t-1}\rightarrow\text{S}_t\leftarrow\text{I}_{t-1}$ is present in the female subcohort with a posterior value of 0.1, but the same structure had a posterior of 0.04 for males and was rejected. Additionally, the male subcohort has the structure $\text{S}_{t-1}\rightarrow\text{S}_t\leftarrow \text{A}_t$ with a posterior of 0.24. Meanwhile, the same structure was rejected in the female subcohort due to having a posterior of 0.04. Finally, While the male subcohort has a dependency across time for F, with the structure $\text{F}_{t-1}\rightarrow\text{F}_{t}$, having a posterior of 0.96, the value of the feature for the female subcohort was 0 across the board, leading to no structure with node F. This complete absence of flow-limited respiratory effort (F) in the female subcohort might be due to the low sensitivity of feature construction to small values.   

Meanwhile, when considering the age-based subcohorts, the key difference between them is the causal relationship between Apnea Hypopnea Events (A) and Snoring (S) in the same window. For the older cohort, the causal direction is from A to S, with the direction reversed for the young. This is due to the different structures admitted in the two cases. The structure $\text{A}_t\rightarrow\text{S}_t\leftarrow\text{S}_{t-1}$ is admitted for the older cohort, with a posterior of 0.35. The same structure had a posterior of 0.14 for the young subcohort, but the structure with a higher posterior of 0.24,  $\text{A}_{t-1} \rightarrow\text{A}_t\leftarrow \text{S}_t$  was admitted instead. 

Finally, when we consider the different subcohorts as well as a time-series causal model trained across the entire cohort, we find that the temporal dependencies between consecutive windows for the same feature are the most prominent. Moreover, the dependency between Apnea Hypopnea Events and Desaturation persisted in all cases. However, other relationships may vary greatly across subcohorts, with some causal relationships reversed and others completely absent, indicating possible differences in sleep dynamics across subcohorts.
\section{Discussion}
We considered the task of modeling the causal dynamics of sleep-disordered breathing using data from a home sleep apnea test (HSAT). Such causal models are necessary for building reliable DDDAS systems for clinical decision support.  We combined the PCMCI$+$ dynamic causal discovery algorithm with bootstrap aggregation based on higher-order structure to learn time-series causal graphs from a small dataset. We compared the structures learned from age- and sex-stratified subcohorts, finding that temporal self-dependencies and apnea-desaturation relationships persist across subcohorts while several substructures differ substantially, including a reversal in the causal direction between apnea and snoring across age groups. 

There are several directions for future work. First, expanding the analysis to include additional features
would be important future work, since HSAT recordings do not capture all physiologically relevant variables, such as sleep stage and arousal. Second, non-stationary causal discovery methods, such as RPCMCI~\cite{saggioro2020reconstructing}, would better account for the fact that sleep dynamics shift throughout the night as the patient cycles through sleep stages. Finally, incorporating additional background knowledge of higher-order structures could yield more accurate models for sleep dynamics.
\bibliographystyle{splncs04}
\bibliography{refs}

\end{document}